\documentclass[11pt]{article}
\usepackage{arabtex}
\usepackage{coling2018}
\usepackage{utf8}
\setcode{utf8}
\usepackage{url}
\usepackage{graphicx} 
\usepackage{tikz}
\usepackage{caption} 
\usepackage[stable]{footmisc}
\date{}

\begin{document}

\title{\textbf{The First Parallel Corpora for Kurdish Sign Language}}

\author{
	\begin{tabular}[t]{c}
		Zina Kamal and	Hossein Hassani\\
		\textnormal{University of Kurdistan Hewl\^er}\\
		\textnormal{Kurdistan Region - Iraq}\\
		{\tt {\{z.kamal3, hosseinh}\}@ukh.edu.krd}
	\end{tabular}
}

\maketitle

	%
	%
	\begin{abstract}
		Kurdish Sign Language (KuSL) is the natural language of the Kurdish Deaf people. We work on automatic translation between spoken Kurdish and KuSL. Sign languages evolve rapidly and follow grammatical rules that differ from spoken languages. Consequently, those differences should be considered during any translation. We proposed an avatar-based automatic translation of Kurdish texts in the Sorani (Central Kurdish) dialect into the Kurdish Sign language. We developed the first parallel corpora for that pair that we use to train a Statistical Machine Translation (SMT) engine. We tested the outcome understandability and evaluated it using the Bilingual Evaluation Understudy (BLEU). Results showed 53.8\% accuracy. Compared to the previous experiments in the field, the result is considerably high. We suspect the reason to be the similarity between the structure of the two pairs. We plan to make the resources publicly available under CC BY-NC-SA 4.0 license on the Kurdish-BLARK (\url{https://kurdishblark.github.io/}).
	\end{abstract}
	%
	%
	\section{Introduction}
	\label{sec:intro}
	
	Sign language is the natural language of the Deaf that uses visual interactions rather than sound. The visual interaction could be manual, non-manual, and finger spelling. This characteristic of sign language adds challenges to its processing. The lack of a writing system is another challenge in this regard. Glosses and Notations are available to overcome these issues.
	
\begin{figure}[ht!]
		\centering
		\includegraphics[scale=0.35]{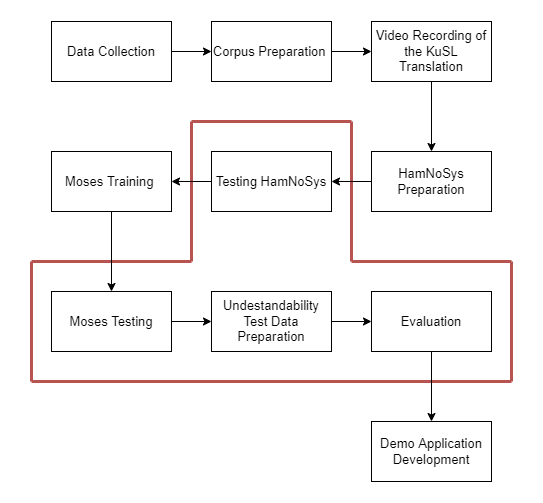}
		\caption{Methodology Flow  }
		\label{archm}
	\end{figure}

	In this research, we develop the first parallel corpora between spoken Kurdish and Kurdish Sign Language (KuSL). Also, we test the understandability of avatars from the Deaf. We trained a statistical Machine Translation (MT) engine (Moses) to translate between spoken Kurdish and KuSL. We evaluate the outcome using BLEU. In order to animate the translated sentences, we prepare the HamNoSys equivalent for all the translated KuSL glosses. To develop the glosses, we record the Deaf signing the sentences of the corpora. Although we do not use the recordings directly in the translation process, we consider them to be highly valuable for future research. We use an avatar to show the corresponding glosses to the translation audience. Figure \ref{archm} shows the method we follow to achieve our objectives.

	The rest of the paper is organized as follows. Section \ref{Sec-KuSL} gives a background about Kurdish and KuSL. Section \ref{Sec-Work} provides a brief insight into related work. Sections \ref{Sec-CorpusPrep} and \ref{Sec-HNS Prep} explain different stages of the data collection process. Section \ref{Sec-Exp}  illustrates the performed experiments. In Section \ref{Sec-Test}, we show the testing approaches we used in the study. Section \ref{Sec-Result} reports the result and discusses the outcomes. Finally, we summarize our conclusion and future work in Section \ref{Sec-Conc}.

	\section{Kurdish Sign Language }
	\label{Sec-KuSL}
	
	Kurdish is an Indo-European language spoken in multiple dialects. It is considered a low-resourced language where work on processing it started evolving only in the last decade \cite{abdulrahman2020using}. The resources for KuSL processing are severely limited. To the best of our knowledge, previous work on KuSL study and Kurdish enhancement to be sign-supported is severely restricted \cite{kamal2020towards}. 
	
	Three cities within the Kurdistan region of Iraq (KRI), Erbil, Suleimany, and Dohuk, currently use KuSL for education. Despite using different spoken dialects/sub-dialects in those cities, their Deaf communities use the same sign language. Work on KuSL development started in early 2000 with the first printed dictionary for Kurdish signs \cite{jepsen2015sign}. Later, in 2014, the printed dictionaries were revised and expanded to support education in special schools for the Deaf. The dictionaries included KuSL signs from several topics, such as Mathematics and Kurdish. Since then, the KRI Deaf schools have been using those dictionaries in their teaching process. However, as sign languages are not entirely standardized, modifying them can be difficult. Having online sign dictionaries minimizes the modification challenges. Our attempt is the first to make online data for KuSL available. 
	
	\section{Related work}
	\label{Sec-Work}
	
	Machine translation (MT) techniques have been used to translate spoken into sign language. \newcite{almasoud2011proposed} proposed a semantics translation system between Arabic texts and Arabic Sign language (ArSL) using a rule-based approach. They built the system based on the ArSL grammatical transformation rules. They also used domain ontology for semantic translation and SignWriting notation production. The semantic translation used the domain ontology to convert texts into SignWriting symbols that the Deaf could read.
	
	\newcite{stoll2018sign} used Neural Machine Translation (NMT) and image generation techniques to develop a system for English to ASL translation. They used video streaming rather than an avatar. Their system can translate sentences from spoken language to ASL by giving the output as video streams. They also used a generative model to produce video sequences using the pose information. They used BLEU to test the system's efficiency and reported 16.34/15.26 BLEU scores on dev/test sets. 
	
	\newcite{su2009improving} developed parallel bilingual corpora to translate Chinese into Taiwanese Sign Language (TSL). They adopted a Synchronous Context Free Grammar (SCFG) to convert spoken Chinese sentences into TSL signing glosses. Statistical Machine Translation (SMT) was used to extract statistical information between the translations by referring to the translation memory. The experiments’ results showed that the translation memory played an essential role in making the SMT more efficient than the traditional SMT, specifically for small bilingual corpora that include long sentences.

	\section{Corpus Preparation}
	\label{Sec-CorpusPrep}
	
	To develop our parallel corpora, We use Kurdish Textbook Corpus (KTC) \cite{abdulrahman2019ktc}, which was developed based on the Kurdish textbooks used in the K-12 schools of the Kurdistan Region of Iraq. We expanded the mentioned corpus to include texts from elementary grades as well. The  selected sentences were in different form areas, such as poetry and short stories, from the Kurdish language textbooks. Table \ref{SentType} shows the number of sentences of each category. The chosen sentences are of various lengths. The entire process prepared glosses for approximately 500 Kurdish sentences.

	\begin{table}[ht!]
		\fontsize{10}{12}\selectfont
		
		\begin{center}
			\caption{Sentence Type}
			
			\begin{tabular}{|c|c|}
				
				\hline
				\textbf{Sentence Source} &  \textbf{Sentence Number} \\ 
				\hline
				Short Story  & 111      \\
				\hline
				Poems  & 105  \\ 
				\hline
				Short Independent Sentences   & 284 \\ 
				\hline
				Total   & 500 \\
				\hline
				
			\end{tabular}
			\label{SentType}
		\end{center}
	\end{table}
	
\begin{figure}[ht!]
		\centering
		\includegraphics[scale=0.6]{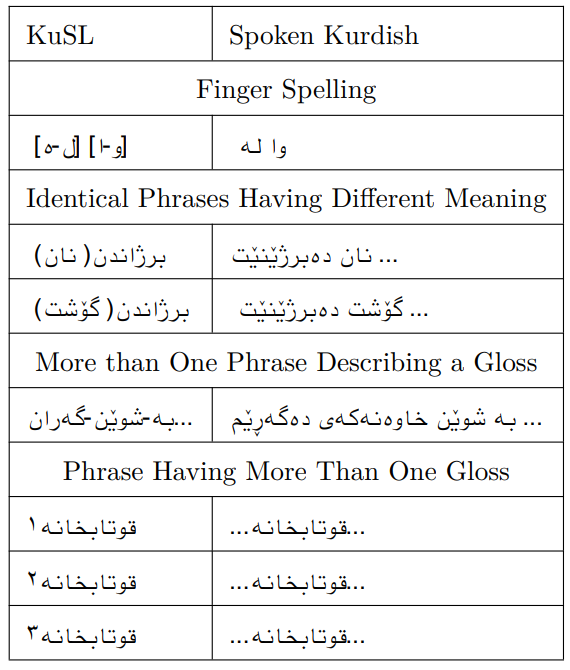}
		\caption{Gloss Identification  }
		\label{gloss}
	\end{figure}

\begin{figure}[ht!]
		\centering
		\includegraphics[scale=0.65]{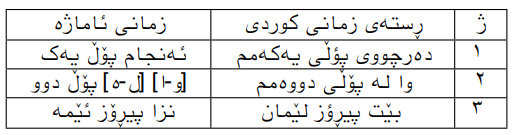}
		\caption{Corpus Sample  }
		\label{crps}
	\end{figure}

	The KuSL interpreter read the sentences to be translated and explained the concept to the Deaf to let them propose suitable signs for each sentence. As a result, the glosses for each sentence was selected by the Deaf, which could increase the understandability of the model's outcome. We aligned the translated Kurdish glosses with the sentences and labeled them sequentially (see Figure \ref{crps} and Table \ref{TraCorpSamp}). Because, currently, no glossing rules exist for KuSL labeling, we adopted \newcite{johnston2008corpus} rules to gloss KuSL as Figure\ref{gloss} displays. The words expected for finger spelling are separated by ``-'' and grouped with square brackets ([]) to be fingerspelled in sequence. For phrases whose meanings are context-dependent, we put the distinguishing word into brackets ``()''. If more than one word is required to describe a gloss, they are separated by ``-'' to form a consolidated gloss. In other cases where a gloss represents more than one sign, we distinguish the glosses by different numbers.

	\renewcommand{\arraystretch}{0.7}
	\begin{table}[ht!]
		\begin{center}
			\caption{Translated Corpus Sample }
			
			\small
			\begin{tabular}{|c|c|c|}
				\hline
				\textbf{No.} & \textbf{ Kurdish Sentence} &    \textbf{KuSL Glosses} \\ 
				\hline
				1 & I passed grade one &  PASS GRADE ONE  \\
				\hline
				2 & I'm in grade two &  GRADE TWO [I-N]\\ 
				\hline
				3 & Congratulations to us & CONGRATS US \\ 
				\hline
				
			\end{tabular}
			\label{TraCorpSamp}
		\end{center}
	\end{table}

	\section{HamNoSys Preparation}
	\label{Sec-HNS Prep}
	
\begin{figure}[ht!]
		\centering
		\fbox{	\includegraphics[scale=0.46]{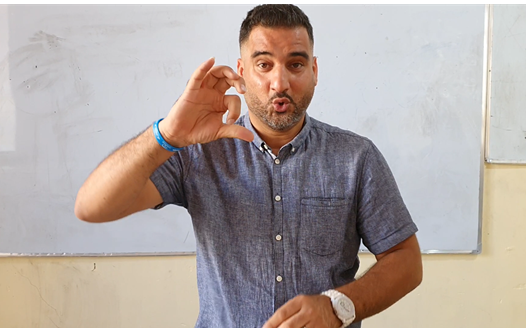}}
		\caption{Physical Sessions }
		\label{vdeoschl}
	\end{figure}

	We created the HamNoSys for each corpus gloss. We recorded the signing of each sentence in the corpus and used them as guidelines during the transformation. We started having face-to-face meetings and recordings at the Deaf school in Erbil city in the Kurdistan Region of Iraq. Figure \ref{vdeoschl} shows a screen from one of the data collection sessions. However, we could not complete the session as planned due to the Kurdistan Region Government (KRG) COVID-19 regulations that suspended the schools in November 2020. We had to stop recording for a while and then resumed it using online platforms (Zoom). At first, the new recording method was challenging considering several parameters such as quality, privacy, and internet connection. Before starting the actual sessions, we performed two test sessions to assess the environment and various techniques to record the videos. Finally, within five months, we could finish the planned sentences. Then, we created the HamNoSys transformation for each gloss. We used the eSign editor to concatenate the HamNoSys segments of the glosses and form the sentences for the avatar in the SiGML player.

	\section{ Experiment}
	\label{Sec-Exp}
	
	We separate the corpus into two sections: 80\% of it for training and 20\% for testing. We use Moses to train our translation pair. Moses is an open-source statistical machine translation engine with three main contributions: support for linguistically motivated factors, decoding for confusion networks, and data format efficiency for language and translation models \cite{koehn2007moses}. Researchers have already used Moses to translate spoken sentences into signing glosses. We follow that approach to evaluate the understandability and the accuracy of SMT for the Deaf and assess the efficiency of Moses using the Bilingual Evaluation Understudy (BLEU) automatic MT evaluator. BLEU is an evaluating algorithm for automatic translation. It evaluates the output of a translation and indicates how close it is to a human translation. 
	
	\section{ Testing and Evalution}
	\label{Sec-Test}
	
	As shown in Figure \ref{archm}, the experiments went through three different test phases: evaluation of the trained language mode using BLEU, testing the well-formedness of the HamNoSys transformation to obtain a more natural avatar signing, and assessing the understandability of the glosses played by the avatar.
	
	We evaluated the well-formedness of the HamNoSys transformation. We prepared the HamNoSys for sets of sentences and played them to the Deaf. The Deaf viewed the avatar playing the target sentences and gave feedback. The process was accomplished online by sending the videos through WhatsApp and receiving feedback as short videos indicating the signs that required modifications. The modification of the HamNoSys sequences based on the mentioned videos improved the accuracy of the transformation, which in turn led the avatar to animate the signs in a way that resembled the natural singing.
	
	We tested the understandability of the translation of a sentence in an avatar played using three testing categories. In the first category, we tested sentences from the training set itself. This checks if the system can produce a correct translation based on the trained sentences. The second category uses new sentences based on the trained ones. It demonstrates if the machine can generate accurate translations for newly constructed sentences. The third category tested the sentences that were not within the trained sentences. We prepared a survey to determine the level of Deaf understandability (partially, fully, or not at all) of the sentences avatar played.
	
	The evaluation of avatar-based models is often carried out by researchers, not the Deaf \cite{kipp2011assessing}, but we chose to engage the target population to assist us in the evaluation process. In doing so, we adapted the focus group method that \newcite{kipp2011assessing} and \newcite{ebling2016building} used in their research. However, we could not apply it due to the social distancing regulations, as our data collection and experimentation took place during the COVID-19 pandemic. We selected four Deaf people from the Kurdish Deaf community who did not participate in the data collection process during the initial stage. The Deaf were of different ages, two nine years old male and female kids and two adults, one nearly 35 and the other 18 years old. To test our prepared dataset, we assigned two Deaf moderators who were involved in the data preparation method. We showed the testers the avatar playing the signs individually without interrupting, and then we asked them to repeat the signs that they understood. When the test of the four Deaf candidates finished, we discussed their understandability with the other Deaf who had already participated in the data collection process and filled the survey accordingly.
	
	\section{Results and Discussion}
	\label{Sec-Result}
	
	The BLEU results showed 53.8\% accuracy. The percentage might be considered high for such a translation. We suspect the similarity level between the two target languages compared when the pairs were from different languages to be the reason for such a high rate. Also, our data was from a restricted context with only a small number of sentences.

	\par
\begin{figure}[ht!]
		\centering
		\fbox{	\includegraphics[scale=0.34]{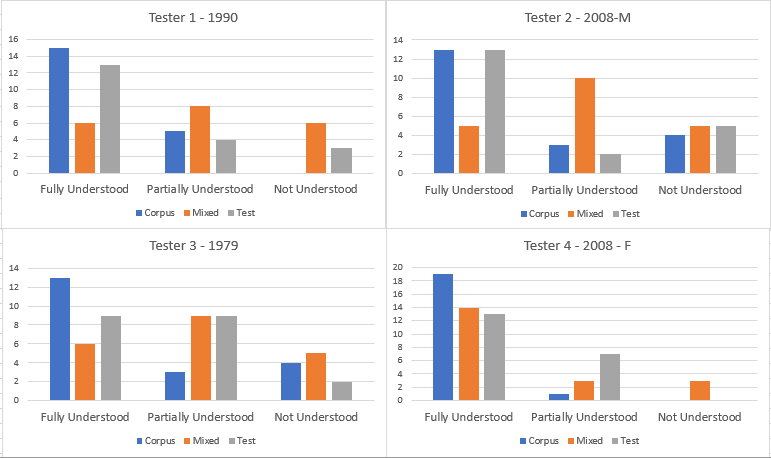}}
		\caption{Understandability Test Outcome}
		\label{allev}
	\end{figure}

 After conducting our test procedures, we analyzed the survey results to evaluate the accuracy of Moses's translation and the understandability of the translated sentences. As in Figure \ref{allev},  the mixed sentences had the highest level of misunderstood and the lowest fully understood sentences. That indicates that even if all the phrases are found inside the language model, Moses still requires more sentences to provide higher accuracy. For instance, providing phrases used in various contexts in the trained model would lead to a more accurate translation. It is also clear from the graph that all the testers could fully understand the sentences of the trained corpus. Although it takes a lot of time and effort, the HamNoSys well-formedness test played an essential role in achieving such a result.

\begin{figure}[ht!]
		\begin{center}
			\fbox{	\includegraphics[scale=0.5]{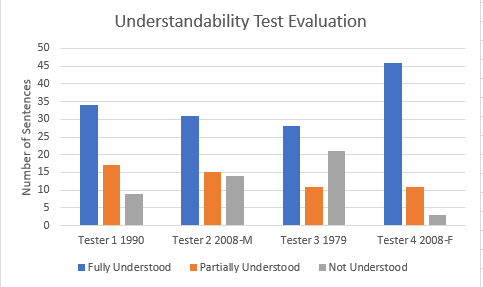}}
		\end{center}
		\caption{Overview of Understandability Evaluation Results}
		\label{evtest}
	\end{figure}

	During the evaluation test, With each passing day, the Deaf were becoming more familiar with the avatar and getting used to it. The young Deaf were more comfortable with the avatar, while the adult Deaf needed more time to interpret the played signs by the avatar. Toward the end of the evaluation sessions, the Deaf were more comfortable with the avatar and had more intention to give feedback and suggest modifications. Figure \ref{evtest} shows the overview of the understandability test outcome

	\section{Conclusion}
	\label{Sec-Conc}
	
	We presented the first parallel corpus between spoken Kurdish and KuSL to train an SMT engine using Moses. We observe the understandability of such translation from the Deaf. We tested its outcome from three different categories. Also, we initiate the first step for having data available online for KuSL.
	In the future, we are interested in expanding our corpora to cover a wider area that could help the understandability of the translation to be improved.
	
	\section{Acknowledgements}
	
	We wholeheartedly acknowledge the assistance of the Deaf community and their kind support during the data collection. Our special gratitude goes to Ms.Mihan Fattah for her kind support during the research. We thank the KuSL interpreter Ms. Samira for her support throughout the data collection process.Also, Ahmad Ajib Ali and Shno Ahmad for the patience they showed during the data collection process. We appreciate all the beautiful-hearted Deaf (Ramsya, Yasin, Lana, and Shayma) who participated in our testing sessions.
	
	\section{Limitations}
	We worked on the first attempt to develop a parallel corpus of Kurdish and KuSL. The main limitation of our work was the lack of KuSL glosses. Having a wider range of sentences would bring better outcomes for statistical approaches. 
	\section{Ethics Statement}
	For all the participating Deaf we required them to sign a ethical consent that included their agreement to publish their videos on the Internet and use them in future scientific works.



	\bibliographystyle{lrec}
	\bibliography{KuSignDataset}

	\end{document}